\documentclass[letterpaper, 10 pt, conference]{ieeeconf} 

\IEEEoverridecommandlockouts

\usepackage{amsmath}
\usepackage{amssymb}
\usepackage{graphicx}
\usepackage{booktabs}
\usepackage{makecell}
\usepackage{bm}
\usepackage{balance}
\usepackage{cite} 
\usepackage{flushend}

\usepackage[dvipsnames]{xcolor}

\definecolor{darkgreen}{rgb}{0,0.6,0}
\definecolor{darkblue}{rgb}{0.1,0.3,0.8}
\definecolor{darkred}{rgb}{0.6,0,0}

\makeatletter
\let\NAT@parse\undefined
\makeatother
\usepackage[bookmarks=true,colorlinks=true,pdfpagemode=UseNone,citecolor=darkgreen,linkcolor=darkgreen,urlcolor=blue]{hyperref}

\bstctlcite{IEEEexample:BSTcontrol}

\title{\LARGE \bf
PACE: Physics Augmentation for Coordinated End-to-end Reinforcement Learning toward Versatile Humanoid Table Tennis
}

\author{Muqun Hu$^{1,*}$, Wenxi Chen$^{1,*}$, Wenjing Li$^{1,*}$, Falak Mandali$^{1,\ddagger}$, Zijian He$^{1,\ddagger}$, Renhong Zhang$^{2,\ddagger}$, \\Praveen Krisna$^{1,\ddagger}$, Katherine Christian$^{1,\S}$, Leo Benaharon$^{1,\S}$, Dizhi Ma$^{3,\S}$, Karthik Ramani$^{1}$, Yan Gu$^{1,\dagger}$
\thanks{\newline $^{1}$ School of Mechanical Engineering, Purdue University, West Lafayette, IN 47907, USA.\newline $^{2}$ College of Engineering, Purdue University, West Lafayette, IN 47907, USA. \newline $^{3}$ School of Electrical and Computer Engineering, Purdue University, West Lafayette, IN 47907, USA. \newline $^{*}$ Equal contribution (Group 1). $^{\ddagger}$ Equal contribution (Group 2). $^{\S}$ Equal contribution (Group 3). \textsuperscript{\dag}Corresponding author. E-mail: yangu@purdue.edu}
}

\begin{document}

\bstctlcite{BSTcontrol}
\maketitle
\thispagestyle{empty}
\pagestyle{empty}

\begin{abstract}

Humanoid table tennis (TT) demands rapid perception, proactive whole-body motion, and agile footwork under strict timing—capabilities that remain difficult for end-to-end control policies. We propose a reinforcement learning (RL) framework that maps ball-position observations directly to whole-body joint commands for both arm striking and leg locomotion, strengthened by predictive signals and dense, physics-guided rewards. A lightweight learned predictor, fed with recent ball positions, estimates future ball states and augments the policy’s observations for proactive decision-making. During training, a physics-based predictor supplies precise future states to construct dense, informative rewards that lead to effective exploration. The resulting policy attains strong performance across varied serve ranges (hit rate$\geq$96\% and success rate$\geq$92\%) in simulations. Ablation studies confirm that both the learned predictor and the predictive reward design are critical for end-to-end learning. Deployed zero-shot on a physical Booster T1 humanoid with 23 revolute joints, the policy produces coordinated lateral and forward–backward footwork with accurate, fast returns, suggesting a practical path toward versatile, competitive humanoid TT. We have open-sourced our RL training code at: \href{https://github.com/purdue-tracelab/TTRL-ICRA2026}{https://github.com/purdue-tracelab/TTRL-ICRA2026}.

\end{abstract}

\section{Introduction}

Humanoid robots hold great promise as general-purpose embodied intelligent agents capable of performing diverse real-world tasks~\cite{gu2025evolution,misenti2025experimental}. Recent advances in learning and control have substantially expanded their physical capability, with impressive demonstrations of walking \cite{radosavovic_real-world_2024,gao2025time,stewart2025adaptive}, jumping \cite{he_asap_2025}, kicking~\cite{tay2026hybrid}, and dancing \cite{ji_exbody2_2024}. Still, most research has focused on scenarios where humanoids operate in free space or interact only with static objects. Reacting to fast-moving objects with versatile, coordinated behaviors remains a fundamental challenge in humanoid whole-body control (WBC).

Table tennis (TT), both a cerebral and physically demanding sport, exemplifies this challenge. Competitive TT matches are characterized by breathtakingly dynamic exchanges that demand rapid eye-hand-leg and whole-body coordination as well as high-level strategies that adapt to an opponent’s strengths and weaknesses. Versatile stroke strategies, which require agile footwork, core stabilization, and precisely coordinated arm swing, play a key role in winning a TT match~\cite{faber2021developing, wong2020biomechanics}. In contrast to some of the previous dynamic locomotion or quasi-static manipulation tasks, robotic TT is distinguished by the need for high-speed perception, control, and high-level physical versatility—the ball needs to be played very precisely in time and space. 

\begin{figure}[t]
    \centering
    \includegraphics[width=0.45\textwidth]{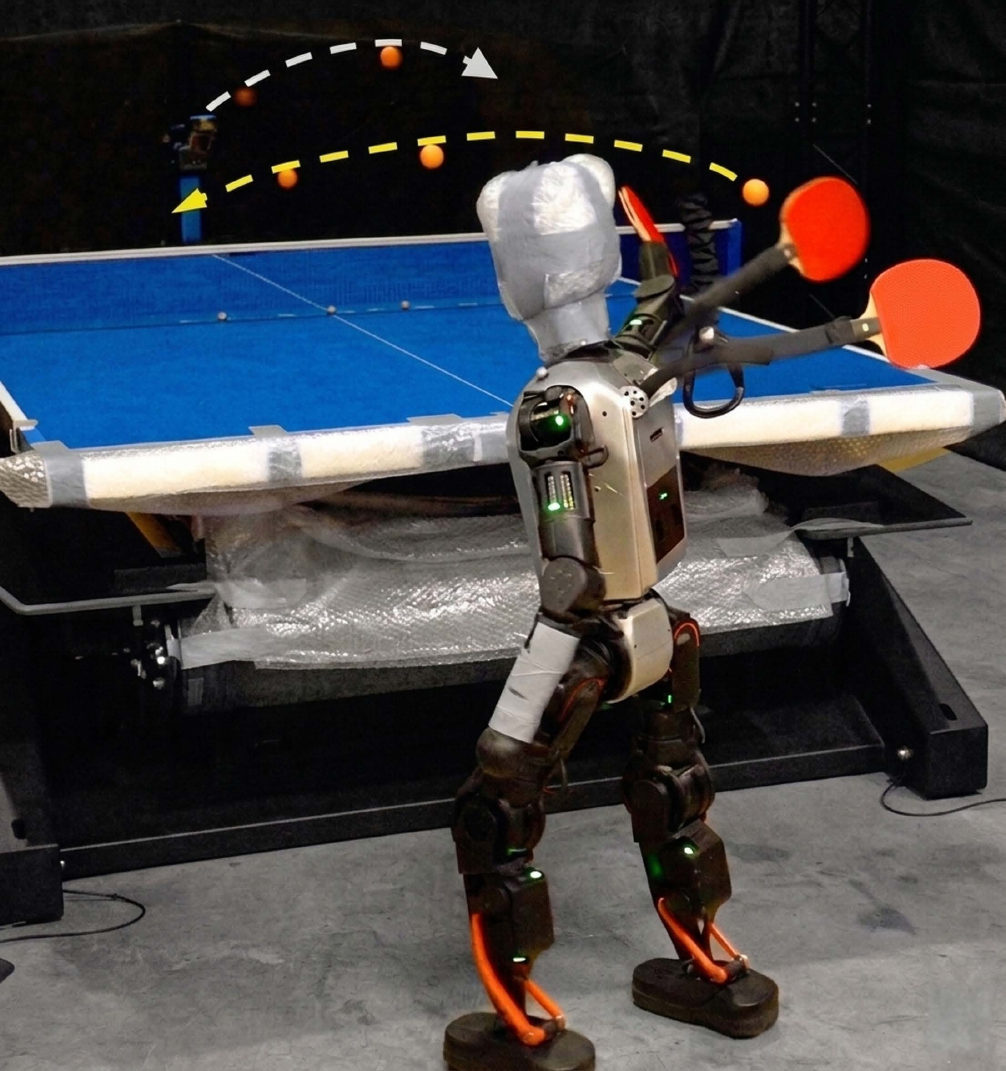}
    \vspace{-0.05 in}
    \caption{The Booster T1 humanoid successfully returns a high-speed ball (6 m/s) from a serving machine. The learned end-to-end whole-body control policy achieves a rapid 0.5 second interception and return, demonstrating coordinated arm-leg movements. Supplementary video: \href{https://youtu.be/CfGj9J0wCEs}{https://youtu.be/CfGj9J0wCEs}.} 
    \label{fig: teaser}
    \vspace{-0.2 in}
\end{figure}

Towards solving these challenges, this paper presents an end-to-end reinforcement learning (RL) framework 
that maps ball-position observations and robot proprioception directly to whole-body reference motions for coordinated arm swing and footwork with a high success rate (Fig.~\ref{fig: teaser}).

\subsection{Related Work}

\subsubsection{Model-based planning and control} Prior work on robotic TT control has been predominantly based on analytical physics-based models, combining ball-trajectory prediction with inverse kinematics for robot motion planning. 

Most approaches adopt the virtual hitting-plane assumption, where a virtual hitting point \cite{RAMANANTSOA1994} is computed from a partial ball trajectory in real-time. With the predicted ball velocity at a given time instant, the target racket pose (i.e., position and orientation) and velocity are then specified, and a robot's motion trajectory is generated to bring the racket to the desired pose at the required time \cite{mulling_biomimetic_2011, Matsushima2005ALA, Miyazaki2006LearningTD, Huang2015LearningOS}.
Yet, the virtual hitting-plane assumption is restrictive.
It reduces TT to a predictable machine task, eliminating the need for highly adaptive footwork and arm-leg coordination, and preventing both humans and humanoids from learning the true variability of play.

Recently, the virtual hitting-plane assumption has been relaxed to enable model-based controllers to realize combat strategies with competitive success rates \cite{ji_model-based_2021} and returning speeds across multiple swing types \cite{nguyen_high_2025}.
Still, most of the existing model-based approaches have been primarily focused on robotic arms with relatively few degrees of freedom (DoFs), making it difficult to directly scale to humanoid whole-body movement, which involves far more DoFs. For example, a humanoid robot has been shown to play TT using impedance control \cite{xiong_impedance_2012}, but its motions are constrained to standing still without agile footwork, thereby limiting the effective hitting range. 

\subsubsection{Learning-based control}

Learning-based methods have made substantial progress toward achieving human-level athletic intelligence in robotic TT \cite{wang_learning_2011,dambrosio_achieving_2025,ding_learning_2022,gao_robotic_2020,tebbe_sample-efficient_2024, wang2024strategyskilllearningphysicsbased, Buchler2022Learning, guist2024safe, guist2023hindsightstatesblendingsim, huang2016jointly}.

Concurrently with this work, \emph{HITTER} \cite{su_hitter_2025} hierarchically integrates an analytical-model-based interception planner with an RL controller for humanoid whole-body TT, yielding the first demonstration with agile footwork and accurate strikes. However, due to its underlying virtual-hitting-plane constraint, the robot primarily exhibits lateral footwork.

Researchers at Google DeepMind have introduced an end-to-end, model-free RL framework that relaxes the virtual hitting-plane assumption and learns a perception-to-action policy on a robot arm \cite{gao_robotic_2020}. The robot arm has six revolute joints, with its base sliding along two orthogonal axes (lateral and forward–backward).
This configuration enlarges the reachable workspace and helps enable diverse emergent strokes.
Subsequent work demonstrates competitive, strategy-aware play \cite{dambrosio_achieving_2025}. While the end-to-end framework works for robotic arms, its effectiveness on legged humanoids remains an open challenge.
Due to the high-dimensional action space and inherent locomotion instability of humanoid robots, the training will struggle with efficient exploration. Also, the training can suffer from low sampling efficiency due to the sparse nature of ball-hitting rewards \cite{su_hitter_2025}. 

\subsection{Contributions}

Toward versatile humanoid TT, this paper presents an end-to-end unified RL framework that jointly coordinates locomotion and arm striking directly from ball-position observations and robot proprioception, without relying on modular decomposition or hierarchical planning. The central technical contribution is a predictor- and physics-augmented end-to-end RL framework that combines a learned future ball-state predictor for proactive whole-body interception with physics-based predictive rewards for dense training supervision, thereby improving learning efficiency and enabling effective striking and coordinated footwork. To the best of our knowledge, this work is among the first unified RL frameworks to generate whole-body motions for successful humanoid TT in both simulation and hardware experiments. The learned policy achieves high success rates at competitive ball speeds across a broad serving range, covering variations in both width and depth.

The remainder of this paper is organized as follows. Section II formulates the problem, followed by the RL design details in Section III. Section IV presents the setup for policy training and experimental validation, while the results are presented and discussed in Section V. Section VI concludes the paper and outlines potential future work.

\section{Problem Formulation}
This section presents the problem formulation for the proposed RL-based humanoid TT control, which aims to simultaneously generate coordinated ball hitting and footwork while ensuring balance.

To provide context for the proposed control approach, it is assumed that a low-level joint controller (e.g., a proportional-derivative (PD) controller) is available to track the reference joint trajectories, and that a perception system is available to provide the real-time pose trajectories of the ball and the robot's trunk.
Thus, the focus of this study is on the generation of the whole-body reference joint trajectories for humanoid TT.

Due to real-world implementation imperfections, such as sensor noise and biases, the robot has access only to partial observation of the true system state. Thus, we formulate the TT control problem as a partially observable Markov decision process (POMDP).

The POMDP is defined by a tuple $\mathcal{M}=\langle\mathcal{S},\mathcal{A},\mathcal{T}, r,\mathcal{O},\gamma\rangle$. 
Here, $\mathcal{S}$, $\mathcal{A}$, and $\mathcal{O}$ are the state, action, and observation spaces, respectively.
$\mathcal{T}: \mathcal{S} \times \mathcal{A} \rightarrow \mathcal{S}$ is the state transition function, and each transition has a reward $r: \mathcal{S} \times \mathcal{A} \rightarrow \mathbb{R}$.
The discount factor of the reward is defined as $\gamma\in[0,1)$.
The full state, partial observation, and action at time $t$ are denoted by $\mathbf{s}_t \in \mathcal{S}$, $\mathbf{o}_t \in \mathcal{O}$, and $\mathbf{a}_t \in \mathcal{A}$, respectively.

The POMDP will be solved by RL that generates the reference joint motions for the robot's upper body and legs.
To capture the temporal behavior of the ball and robot motions, we define the observation history at time $t$ over the past $H$ measurements as $\mathbf{o}_t^H=\begin{bmatrix}
 \mathbf{o}_{t}, \mathbf{o}_{t-1}, \ldots, \mathbf{o}_{t-H+1}
\end{bmatrix}$.

The objective of the RL design is to train a policy $\pi_{\theta}$ with parameter $\theta$ that maximizes the expected total discounted return. The policy learns to map the sequence of observations $\mathbf{o}_t^H$ to a probability distribution over actions $\pi_{\theta}$. The optimization objective is defined as $J(\pi_{\theta})=\mathbb{E} \left [ \sum_{t=0}^\infty \gamma^t r(\boldsymbol{s}_t, \boldsymbol{a}_t) \right ]$. By optimizing this objective, the agent learns a control strategy capable of effective ball return, flexible and dynamic footwork, and balance maintenance.

\begin{figure*}[t]
    \centering
    \includegraphics[width=\textwidth]{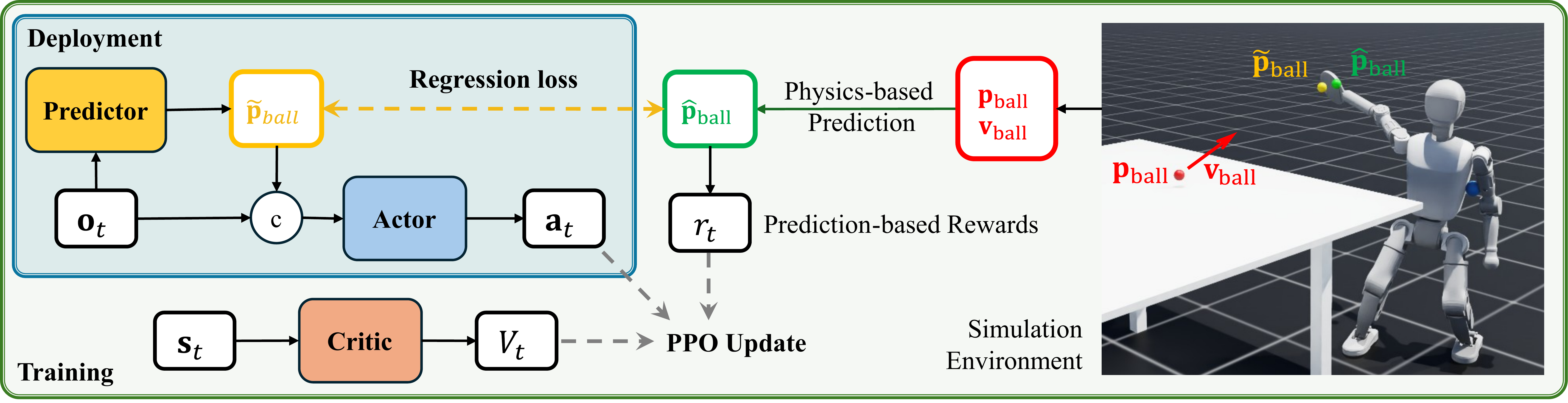}
    \vspace{-0.7cm}
    \caption{Overview of the training pipeline. A learnable predictor anticipates the future desired hitting position of an incoming ball, $\tilde{\mathbf{p}}_{ball}$. Physics-based simulation provides a dynamics-model-based prediction, $\hat{\mathbf{p}}_{ball}$, which serves both as ground truth for training the predictor and as the basis for constructing dense, continuous rewards (illustrated in Fig.~\ref{fig: prediction-based reward}).     
    }
    \label{fig:training_pipeline}
    \vspace{0.5cm}
\end{figure*}

\begin{figure}[t]
    \centering
    \includegraphics[width=0.47\textwidth]{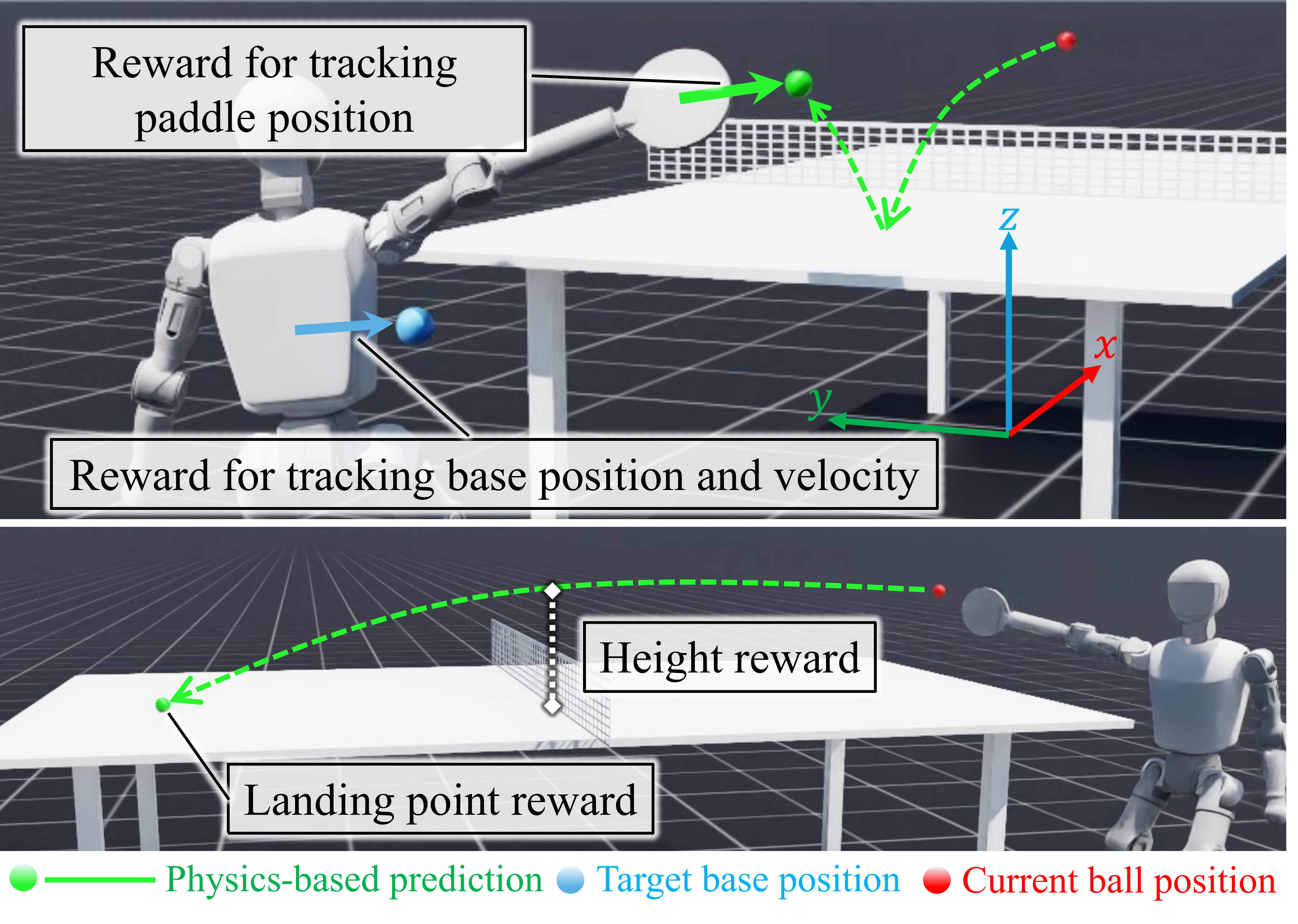}
    \vspace{-0.2cm}
    \caption{Prediction-based reward design for ball-hitting motions. Using the ball physics in simulation, we anticipate the ball’s trajectory (green dashed line). From this, we define a \textit{hit-guidance reward} that encourages the robot to move proactively to intercept the ball (as visualized by the solid green and blue arrows), and a \textit{return-guidance reward} that scores each strike based on the predicted landing point and the ball’s height at the net. Together, these continuous rewards enable the robot to refine its strikes and achieve successful returns.}
    \label{fig: prediction-based reward}
\end{figure}

\section{Methodology}

This section introduces the design of the proposed end-to-end RL framework for humanoid whole-body TT.

A main challenge in training a unified whole-body RL policy for humanoid robotic TT is the need to act proactively on a fast-moving incoming ball while learning from sparse and delayed successes.
To address this challenge, we introduce a unified framework integrating a ball-hitting position predictor and an asymmetric actor-critic RL architecture \cite{pinto_asymmetric_2017} for effective training with efficient and flexible skills exploration, as illustrated in Fig.~\ref{fig:training_pipeline}. At the core of the proposed framework are the design of the predictor-augmented actor policy (subsection A) and the predictive, dense reward design (subsection B). The design of the observation and action spaces for the whole RL framework is described in subsection C.

\subsection{Predictor-augmented Actor Policy}


In a TT game, the ball crosses the table in less than 0.5 seconds, leaving the robot with limited time to react. Successfully intercepting the ball requires the robot to move proactively rather than waiting until the ball crosses the net or reaches the player's table.
Hence, ball motion prediction can benefit robot TT control. 

While the future position of the ball is implicitly encoded in its past trajectory, we found it challenging for a single policy to anticipate the ball's fast motion effectively while generating whole-body reference motions. 
Thus, we incorporate the ball trajectory predictor in the proposed RL pipeline, as shown in Fig.~\ref{fig:training_pipeline} and explained next.

\subsubsection{Predictor training} During policy update, we train the prediction network online using simulated ball trajectories until convergence. The predictor receives five observation histories of ball position as input, and outputs a predicted post-bounce apex of the ball, denoted as $\tilde{\textbf{p}}_{\mathrm{ball}}$.
We use the ball's physics, modeled based on IsaacLab simulation and augmented with ball aerodynamics in Section \ref{sec:aero}, to compute the ground truth post-bounce apex, denoted as $\hat{\textbf{p}}_{\mathrm{ball}}$. The network is optimized with a root-mean-square-error-based regression loss between $\tilde{\textbf{p}}_{\mathrm{ball}}$ and $\hat{\textbf{p}}_{\mathrm{ball}}$. To improve robustness against perception noise and latency in deployment, Gaussian noise is injected into the ball trajectories used in training.

\subsubsection{Augmented actor observation}

To enable the robot to move proactively before the ball arrives,
a target base/trunk shift command, denoted as $\Delta\tilde{\mathbf{\textbf{p}}}_{\mathrm{base},xy}\in\mathbb{R}^2$,
is generated using the output of the predictor, $\tilde{\textbf{p}}_{\mathrm{ball}}\in\mathbb{R}^3$.
Here, $(\cdot)_{xy} \in \mathbb{R}^2$ represents the $x$- and $y$-components of a vector $(\cdot) \in \mathbb{R}^3$, which is expressed in the world frame with its three axes illustrated in Fig.~\ref{fig: prediction-based reward}.

This target base shift $\Delta\tilde{\mathbf{\textbf{p}}}_{\mathrm{base},xy}$ indicates the difference between the current and the desired positions of the base projected on the horizontal plane, which the robot needs to minimize before ball strike.
Accordingly, we design the shift as
$\Delta\tilde{\mathbf{\textbf{p}}}_{\mathrm{base},xy}=\tilde{\textbf{p}}_{\mathrm{ball},xy}-\textbf{p}_{\mathrm{base},xy}$.
Here $\textbf{p}_{\mathrm{base},xy}$ is the current base position projected on the horizontal plane.

To encourage proactive movement before ball strike,
both
$\Delta\tilde{\mathbf{\textbf{p}}}_{\mathrm{base},xy}$ and $\tilde{\textbf{p}}_{\mathrm{ball}}$ are concatenated with other observations and provided to the actor.
In contrast, the critic receives the corresponding quantities (i.e., $\Delta\hat{\mathbf{\textbf{p}}}_{\mathrm{base},xy}$) derived from physics-based prediction, as shown in Table~\ref{tab:obs_spaces}. These physics-based values are more precise but only available in simulation, making them suitable as privileged information for training rather than deployment.

\begin{table}
\centering
\caption{Observation spaces for the policy (actor) and critic. The critic receives additional privileged information in training.}
\begin{tabular}{lcc}
\toprule
\textbf{Observation} & \textbf{Actor} & \textbf{Critic} \\
\midrule
Base angular velocity ($\bm{\omega}_{\mathrm{base}}\in\mathbb{R}^3$) & \checkmark & \checkmark \\
Base position ($\mathbf{\textbf{p}}_{\mathrm{base}}\in\mathbb{R}^3$) & \checkmark & \checkmark \\
Projected gravity vector ($\mathbf{g}_{\mathrm{base}}\in\mathbb{R}^3$) & \checkmark & \checkmark \\
Joint positions ($\mathbf{q}\in\mathbb{R}^{21}$) & \checkmark & \checkmark \\
Joint velocities ($\dot{\mathbf{q}}\in\mathbb{R}^{21}$)& \checkmark & \checkmark \\
Previous action ($\mathbf{a}_{\mathrm{last}}\in\mathbb{R}^{21}$)& \checkmark & \checkmark \\
Ball position (${\mathbf{\textbf{p}}}_{\mathrm{ ball}}\in\mathbb{R}^3$) & \checkmark & \checkmark \\
Robot heading vector ($\mathbf{e}_{xy}\in\mathbb{R}^2$) & \checkmark & \checkmark \\
Predictor output (${\tilde{\mathbf{\textbf{p}}}}_{\mathrm{ball}}\in\mathbb{R}^3$) & \checkmark & -- \\
Predictor-based robot shift ($\Delta\tilde{\mathbf{\textbf{p}}}_{\mathrm{base},xy}\in\mathbb{R}^2$) & \checkmark & -- \\
Physics-based prediction (${\hat{\mathbf{\textbf{p}}}}_{\mathrm{ball}}\in\mathbb{R}^3$) & -- & \checkmark \\
Physics-based robot shift ($\Delta\hat{\mathbf{\textbf{p}}}_{\mathrm{base},xy}\in\mathbb{R}^2$) & -- & \checkmark \\
Ball linear velocity (${\mathbf{\textbf{v}}}_{\mathrm{ball }}\in\mathbb{R}^3$) & -- & \checkmark \\
End-effector position (${\mathbf{\textbf{p}}}_{\mathrm{ee }}\in\mathbb{R}^3$) & -- & \checkmark \\
Time for ball to arrive ($t_{\mathrm{arrive}}\in\mathbb{R}$) & -- & \checkmark \\
Serve progress ($t_{\mathrm{serve}}/t_{\mathrm{serve}, \mathrm{max}}\in [0,1]$) & -- & \checkmark \\
Episode progress ($t_{\mathrm{episode}}/t_{\mathrm{episode},\mathrm{max}}\in[0,1]$) & -- & \checkmark \\
Ball touching own table  ($b_{\mathrm{own table}} \in \{0,1\}$) & -- & \checkmark \\
Ball touching paddle ($b_{\mathrm{paddle}} \in \{0,1\}$)  & -- & \checkmark \\
\bottomrule
\end{tabular}
\label{tab:obs_spaces}
\end{table}

\subsection{Prediction-based Reward Design}
Binary metrics such as a successful hit or return, which encourage arm swing, are too sparse to effectively train an RL agent to play TT. These simple rewards do not provide enough guidance for training the robot's locomotion and arm swing policy, which is why we could not learn a functional WBC policy for humanoid TT using them alone. Instead, we use the ball's physics to predict its future trajectory in order to construct dense, continuous reward functions. Two key reward designs are critical for our learning approach, as illustrated in Fig.~\ref{fig: prediction-based reward}.

\subsubsection{Reaching reward}
During policy update, the target ball hitting position $\hat{\textbf{p}}_{\mathrm{ball}}$ is chosen as the ball's highest point after a bounce, and is predicted based on the ball's physics, as explained earlier. To guide the robot toward this target, we formulate two positional reward terms.

The first punishes the distance between the end effector and the target hitting point ($||{\textbf{p}}_{\mathrm{ee}}-\hat{{\textbf{p}}}_{\mathrm{ball}}||$), 
where $\textbf{p}_{\mathrm{ee}}$ is the end-effector position.
The second penalizes the difference between the robot's target and actual base positions on the $x$-$y$ plane (i.e., $||{\textbf{p}}_{\mathrm{base},xy}+\textbf{p}_{\mathrm{arm},xy}-\hat{\textbf{p}}_{\mathrm{ball},xy}||$).
Here the arm's position relative to the base, $\textbf{p}_{\mathrm{arm}}$, is added to account for the feasible range of the robot's arm reach, positioning the robot's base in a flexible range where the arm can reach the target hitting point. 

Additionally, we generate a pseudo-velocity command for the robot to follow. We penalize the robot for any mismatch between its base velocity and this command (i.e., $||{\textbf{v}} _{\text{base}} -\hat{\textbf{v}}_{\mathrm{ball}} || $), where the target velocity ${\hat{\textbf{v}}}_{\mathrm{ball}}$  is proportional to the tracking error of base position.

We clamp the reward once the position or velocity error falls within a tolerance range around the target, and disable the prediction-based reward near the end of the ball’s flight. This design ensures that the policy retains flexibility to explore effective striking strategies not constrained to hitting the ball exactly at the apex, rather than overfitting to precise position tracking.

\subsubsection{Returning reward}
For the ball to successfully be returned, it must land on the opponent's side of the table and not be blocked by the net. We consider only the case where the ball passes over the net. The success event occurs some time after the ball hitting behavior, after which no control of the ball is possible. To provide immediate feedback after each hit, we use simulation physics to predict the ball’s landing position $(\hat{x}_{\mathrm{ball}},\hat{y}_{\mathrm{ball}})$, and penalize its distance from the desired landing position on the opponent’s side of the table. In addition, we predict the ball’s height $\hat{z}_{\mathrm{ball}}$ as it crosses the net plane and design a reward that encourages the trajectory to pass the net at a specified margin. 

Together, these reward terms provide continuous guidance to the policy, unlike sparse binary success signals, enabling the robot to learn consistent ball-return behaviors. Since they are computed from physics-based predictions, the rewards are available immediately at the moment of contact, rather than only after the ball lands or crosses the net. This design improves learning efficiency by removing the dependence on late outcome signals.

\subsection{Design of Observation and Action Space}

\subsubsection{Observation space design}

To provide the policy with temporal information, the input of the actor network consists of the observation history $\mathbf{o}_{t}^{H}$ with $H = 5$. The actor's observation, detailed in Table \ref{tab:obs_spaces}, is a comprehensive concatenation of the robot's state and its environment. Specifically, it includes proprioceptive information, such as joint positions and velocities, base angular velocity, and projected gravity vector. It also contains external perception information from a motion-capture system, including the robot base position, heading direction, and ball position in the global world frame, along with the robot's previous action, predicted ball apex, and robot base shift. 

The critic network receives this same observation without injected noise, along with additional privileged information. This privileged state consists of ground-truth simulation data that is unavailable on the physical robot but is valuable for learning an accurate value function $V_t$.

\subsubsection{Action space design}
The action space is designed for the policy to learn residual displacements from a stable nominal pose, which promotes safer and more efficient learning. The policy action, $\mathbf{a}_t \in \mathbb{R}^{21}$ is combined with nominal joint values $\mathbf{q}^{\text{nom}}_{t}$ to yield target joint positions $\mathbf{q}^{\text{target}}_{t}$ as $\mathbf{q}^{\text{target}}_{t} = \mathbf{q}^{\text{nom}}_{t} + \alpha \mathbf{a}_{t}$, where $\alpha$ is the action scaling factor.
Here, the nominal joint configuration $\mathbf{q}^{\text{nom}}_t$ corresponds to the robot's standing pose with right-hand raised. $\mathbf{q}^{\text{target}}_{t}$ is tracked by a low-level joint controller (e.g., PD controller).

\section{Training and Experimental Setup}

This section describes the setup for the policy training and experimental assessment.

\subsection{Simulation Setup for Training}

The training environment for humanoid TT is built on LeggedLab~\cite{LeggedLab}, an RL benchmark for humanoid locomotion developed on the IsaacLab platform~\cite{mittal2023orbit}.
The training design is explained next, including details for minimizing the sim-to-real (Sim2Real) gap. 

\subsubsection{Domain randomization} To support Sim2Real transfer, domain randomization is applied to the robot's physical properties (e.g., mass, friction, and restitution). Also, noise and delays are injected into all perception signals, with noise levels matched to the magnitude of real-world sensing errors.

\subsubsection{Air drag modeling}
\label{sec:aero}
To capture aerodynamic effects beyond gravity, which is already handled by the stock IsaacLab, we implement a custom external-force module to account for aerodynamic effects on the TT ball's motion. 
The module was calibrated using 22 real-world pre-bounce trajectories with near-zero spin, measured with the motion capture system: we subtracted gravity from the measured accelerations and fit a quadratic-in-speed drag model \cite{anderson2011ebook} via least squares to obtain a single constant drag coefficient. 

During simulation, at every physics step, the module computes a drag force in a direction opposite to the ball’s instantaneous velocity, with magnitude proportional to the square of its speed, and applies it at the ball’s center of mass in the world frame. Since both the identification and our operating regime involve negligible spin, aerodynamic lift due to Magnus effect is assumed to be negligible. 

\subsubsection{Training episode design} Each training episode begins with a ball initialized near the center of the opponent’s side of the table, approximating the ball-serving machine used during deployment. The ball is served with a random initial velocity, covering both long and short serves, with randomized trajectories spanning a wide range of placements on the forehand side. An episode consists of up to five consecutive serves, but it terminates early if the robot falls. At the start of each new episode, the robot is placed at a random position to encourage exploration of diverse movement directions.

\subsubsection{Actor and critic networks}
The actor and critic networks are multilayer perceptrons (MLP), both with a hidden-layer dimension of [512, 512, 128]. The actor and critic networks are optimized using Proximal Policy Optimization (PPO) \cite{schulman_proximal_2017} with 4096 parallel environments (Fig.~\ref{fig:training_pipeline}). All computations are performed using workstations equipped with NVIDIA RTX 4090 GPUs.

\subsection{Hardware Setup for Real-world Validation}

\subsubsection{TT table}

We use a standard TT table measuring 2.74 m$\times$1.525 m$\times$0.76 m.
The coordinate frame used to express the ball position and the robot's trunk pose is defined as shown in Fig.~\ref{fig: prediction-based reward}. 
\subsubsection{Humanoid robot}
Booster's T1 robot is used considering its robust mobility capabilities. 
The robot is 1.2 m tall and weighs 30 kg. 
It has a total DoF of 23; 4 DoFs in each arm, 6 DoFs in each leg, 1 DoF at the waist, and 2 DoFs for the head (unused).

\subsubsection{Hardware customization} To mount a TT paddle, T1's forearm is replaced with a 3D printed PLA tube mounted at the elbow joint. The TT paddle slides into a friction-fit sleeve placed inside the tube, and is locked at the desired orientation. The tube is sized to imitate the placement of a TT paddle relative to a human player’s arm.  
Note that this mechanical modification does not alter the number of DoFs of the hitting arm.

\subsubsection{Motion sensing}
Vicon's Vero motion-capture system is used to sense the pose of T1's trunk, the table, and the ball, with a tracking accuracy of 5 mm at 150 Hz. Fifteen near-infrared cameras track reflective spherical markers on the robot and the table. Six small pieces of retro-reflective tape are attached to a TT ball in a symmetric pattern, and are sensed by the Vicon system as a cluster of markers.

\section{Results and Discussion}

This section reports the results and analysis from simulations and experiments.

\begin{table}[t]
\caption{Evaluation results for different serve ranges in simulation.}
\label{table: performance in simulation}
\centering
\begin{tabular}{lcccc}
\hline
\textbf{Serve Type} & \textbf{Init. Forward Vel.} & \textbf{Hit} & \textbf{Success} & \textbf{Total Serves} \\
\hline
Long   & $(-6.5,\,-6.2)$ m/s & 96.1\% & 92.3\% & 42181 \\
Mid-long & $(-6.2,\,-5.7)$ m/s& 97.0\% & 95.0\% & 41390 \\
Short  & $(-5.7,\,-5.2)$ m/s& 99.3\% & 94.8\% & 92601 \\
Mixed  & $(-6.5,\,-5.2)$ m/s& 97.6\% & 94.1\% & 88016 \\
\hline
\end{tabular}
\end{table}

\subsection{Training Results from Simulations}
\label{result: training result}

\subsubsection{Performance metrics} 
For both simulations and experiments, two main TT-related metrics are considered: the hit rate and success rate. A hit is defined as intercepting the ball with the paddle, while a success corresponds to making a valid return to the opponent's table.

\subsubsection{High success rates for different serves} 

Table~\ref{table: performance in simulation} summarizes the final simulation performance of our policy
under a large number of trials using parallel environments, with each policy instance playing continuously. In total, we report performance across more than 250,000 serves in simulation.

For evaluation, the ball’s initial speed is fast in the forward direction, ranging from 5.2 m/s to 6.5 m/s.
Its variations categorize the serves into long, short, mid-long, and mixed ranges, as specified in the table.
The lateral velocity is sampled from $(-0.6,0.2)$ m/s to focus on forehand strikes. The initial vertical speed is set between $(1.5,1.9)$ m/s.

Unlike approaches that rely on a predefined virtual hitting-plane, our formulation enables the policy to handle both long and short serves. In the mixed-range setting, where long and short serves are randomly interleaved, the policy must drive the robot to actively move forward and backward to reach effective base positions. Given the limited DoFs of the robot’s forearm, successful returns place a heavier burden on precise locomotion compared to robots whose arms have higher DoFs. Still, the end-to-end policy demonstrates strong performance, achieving a return success rate of about 94\% under mixed serves.

\subsubsection{Diverse strike and footwork strategies} 
\begin{figure}[t]
    \centering
    \includegraphics[width=0.45\textwidth]{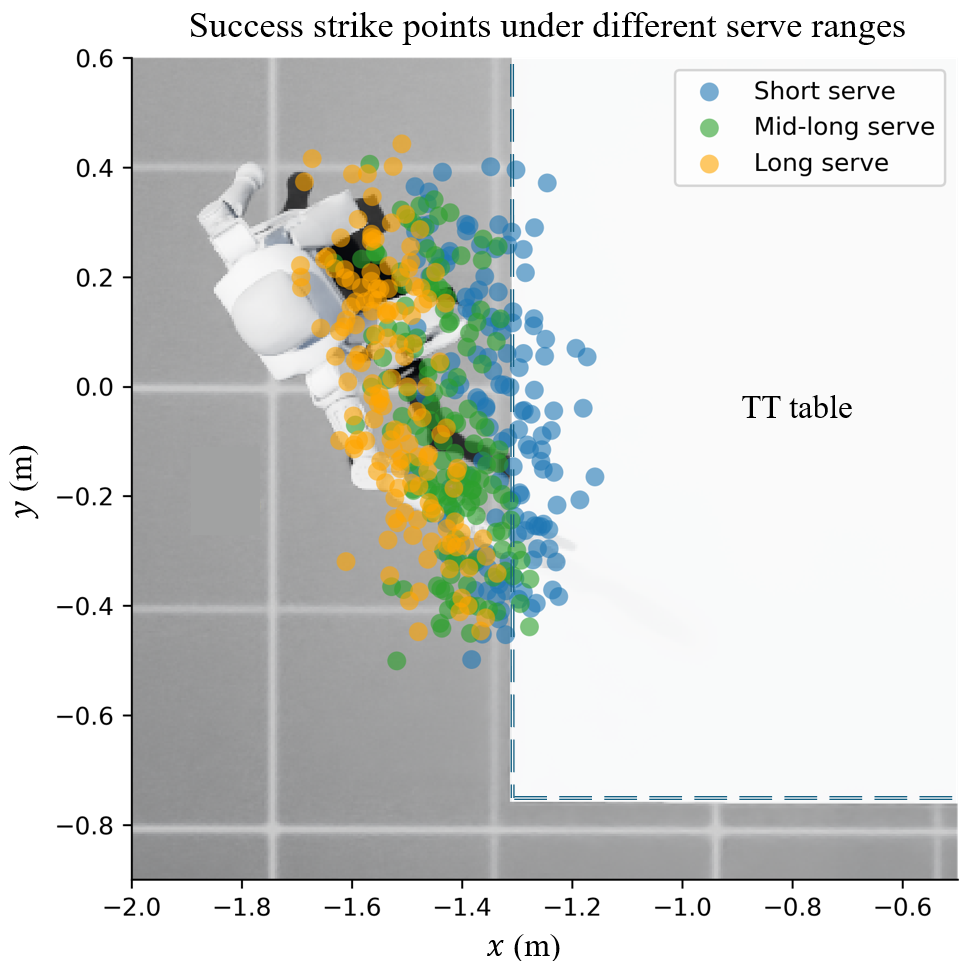}
    \caption{Success strike positions under different serve ranges. Each dot marks the paddle contact point of a successful return. Short and long serves are distinguished by their initial velocity along the $x$-axis of the TT table, consistent with the specifications in Table~\ref{table: performance in simulation}.}
    \label{fig: successful strikes ranges}
\end{figure}

We further visualize the successful strike points for different serve ranges in Fig.~\ref{fig: successful strikes ranges}. The distribution of strike locations shows that the policy adapts its hitting strategy depending on the incoming serve. For short serves, the strikes tend to occur closer to the table, requiring the robot to step forward rapidly, while long serves are intercepted farther back, often involving coordinated backward footwork to maintain balance and reach. This variation highlights that the learned policy does not limit its strikes to a virtual hitting-plane; instead, it leverages whole-body coordination, including locomotion, torso adjustments, and arm swing, to achieve effective returns at diverse locations. These results indicate that the policy has acquired versatile strike behaviors that generalize across various ball trajectories.

\subsubsection{Training of predictor}
We employ a lightweight MLP with two hidden layers of size [64, 64]. Ball trajectories are collected in simulation, and physics-based prediction is used to compute the apex point after the first bounce as ground truth. Training with 20 updates of 4,096 environments and 24 steps per update, which are equivalent to approximately 10 hours of ball trajectories, the predictor achieves an average root-mean-square error of about 3 cm. As shown in our results, this level of accuracy is sufficient to guide the policy and enable consistently high return success rates.

\subsection{Ablation Studies}
\label{result: training ablation}

\begin{figure}[ht]
    \centering
    \includegraphics[width=0.5\textwidth]{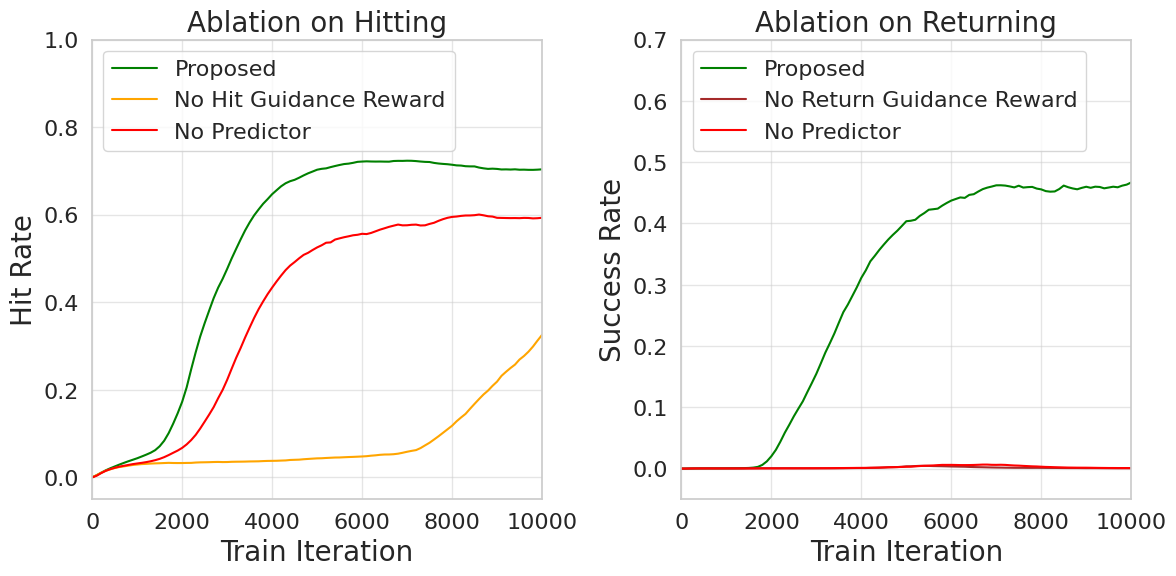}
    \vspace{-0.2 in}
    \caption{Ablation study of the key components of the proposed framework based on training results. Including both the predictor in the actor policy and prediction-based reward designs prove to be crucial for learning successful strikes. Note that the metrics during training are lower than those during testing shown in Table~\ref{table: performance in simulation}. This is due to more frequent environment resets during training for efficient exploration; the immediate post-reset stabilization during training generally leads to missing the first serve.}
    \label{fig: ablation}
\end{figure}

Figure~\ref{fig: ablation} presents the ablation study of the proposed prediction-augmented policy and reward design. In the \textit{no predictor} setting, we remove the future ball position $\tilde{\textbf{p}}_{ball}$ and the prediction-based base shift $\Delta\tilde{\mathbf{\textbf{p}}}_{{base},xy}$ from the actor’s observations, which leads to a sharp drop in hitting performance. Without the predictor, the policy can still move to intercept the ball, but achieves almost no successful returns, as it fails to establish a stable stance or execute an effective strike. This highlights the necessity of the predictor for enabling proactive behavior.

For the prediction-based reward design, the ablation study shows that the \textit{hit-guidance reward} allows the policy to achieve a high hit rate within the first 4000 updates. In contrast, removing this reward leads to less effective exploration, as evidenced by the slower increase in the hit rate in the left learning curve. The \textit{return-guidance reward} is likewise essential for learning effective strike motions. When relying only on sparse success feedback, whether the ball lands on the opponent’s side, the policy never learns the appropriate returning velocity or orientation.

\subsection{Deployment (Sim2Real)}
\label{result: deployment}

We deployed the learned policy zero-shot on the T1 humanoid. A ball-serving machine launched 31 balls from the table center, with initial velocities randomly sampled from $([-5.5, -6.5], [0.5,-0.2],[1.65,1.75])$ m/s, which is within the training distribution. The robot intercepted 29 balls and made 19 valid returns, yielding a hit rate of 93.5\% and a success rate of 61.3\%. 
The Sim2Real gap leading to the decrease, compared to simulation testing, in return success rate  can be attributed to several factors, including the difference between the simulated serial-ankle actuation model and the physical T1's parallel-link ankle mechanism. Although existing framework \cite{wang2025boostergymendtoendreinforcement} converts action for serial linkages to parallel actuation for locomotion policies well, TT playing requires highly dynamic upper and lower-body coordination for precise ball return. A small tracking error in the ankle leads to pronounced end-effector position error. Other contributing factors include unmodeled motor dynamics, as well as imperfectly simulated contact dynamics. 

\begin{figure*}[t]
    \centering
    \includegraphics[width=\textwidth]{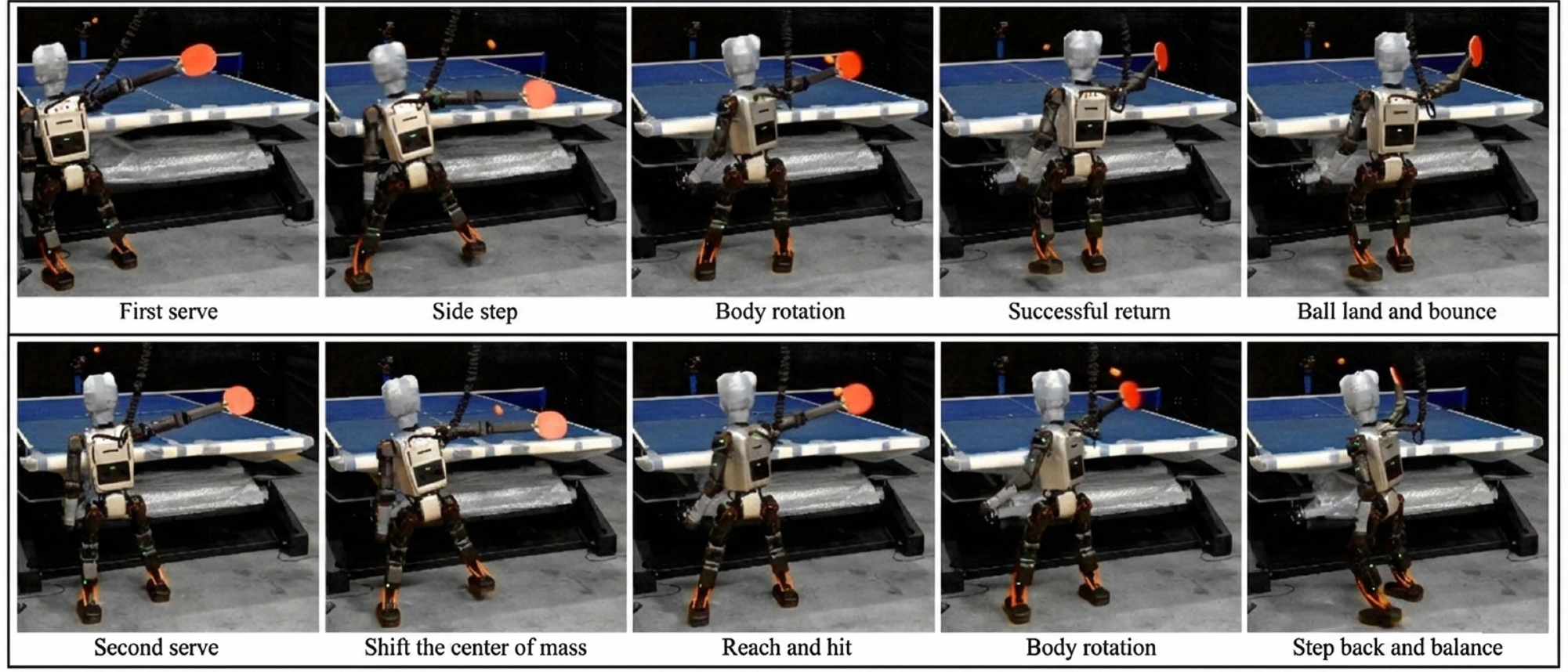}
    \vspace{-0.6cm}
    \caption{Time-lapse sequences of two consecutive returns by T1. On the top row, the arm was predominantly utilized to hit the first ball. On the bottom row, in contrast, the entire trunk rotates to hit the second ball.}
    \label{fig:timelapse}
    \vspace{0.5cm}
\end{figure*}

\begin{figure}[t]
    \centering
    \includegraphics[width=0.47\textwidth]{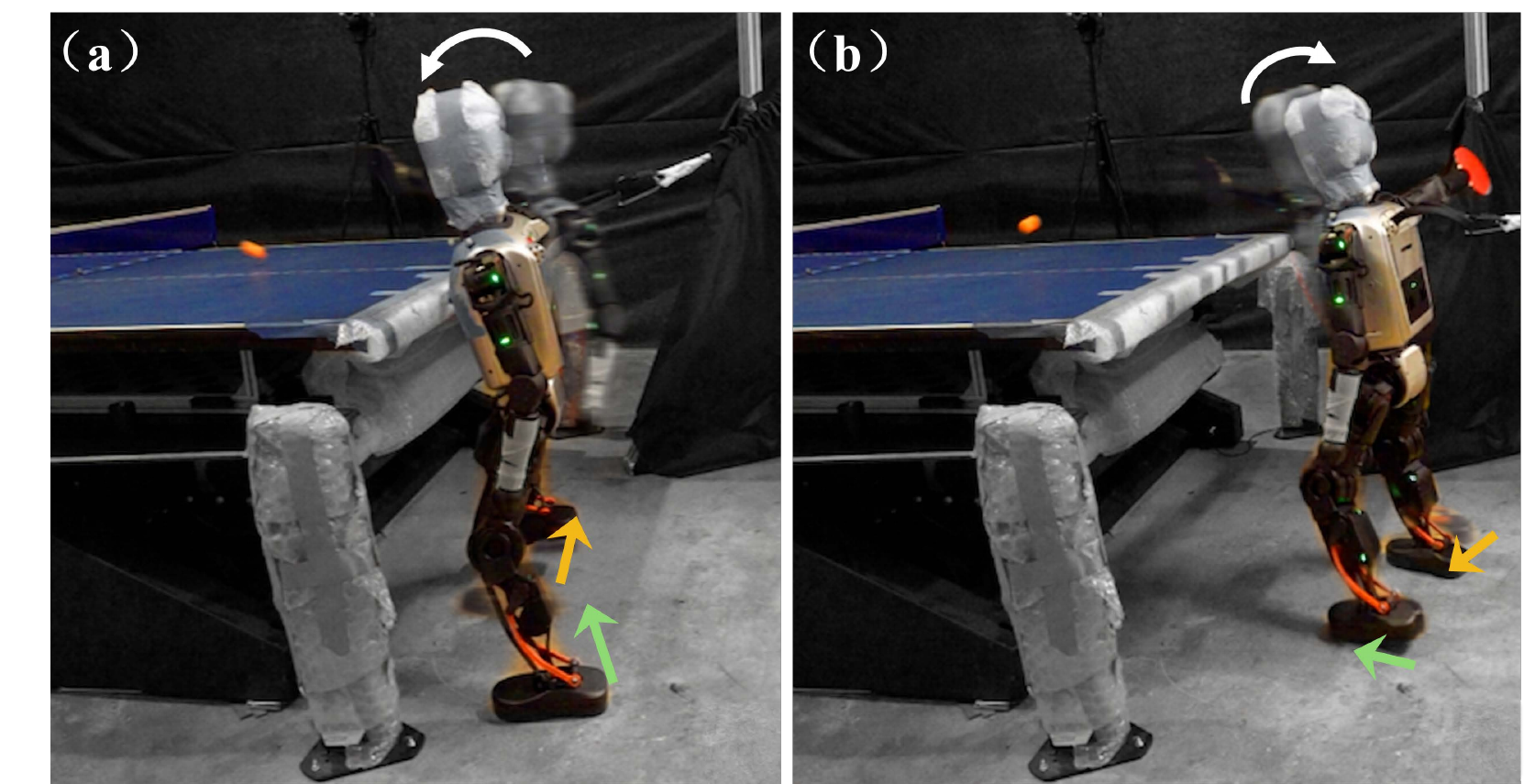}
    \caption{Dynamic two-dimensional footwork before ball strike: (a) lateral and (b) forward-backward movement. The green, yellow, and white arrows indicate the movement directions of the robot's left and right feet, and the trunk.}
    \label{fig: locomotion}
\end{figure}

Consistent with simulation, the policy produced coordinated upper- and lower-body movement. Versatile lateral and forward-backward footwork emerged in response to serves with diverse directions and speeds. 
A time lapse of T1 returning two consecutive balls is shown in Fig.~\ref{fig:timelapse}. When responding to incoming balls in different directions, the robot exhibited versatile behavior during agile locomotion, balancing, and strike-recovery phases.
Figure~\ref{fig: locomotion} contrasts two representative returns with markedly different trunk poses and foot placements, illustrating the policy’s posture and stance adaptation to incoming-ball conditions.

Furthermore, the policy returned the ball at competitive speeds. Across the 19 successful returns, the mean outbound ball speed immediately post-impact was 6.9 m/s, greater than the initial incoming ball speed.

\section{Conclusion and Future Work}
This paper has introduced a unified end-to-end reinforcement learning (RL) framework for humanoid table tennis (TT) control that enables rapid reaction to fast-moving objects with highly coordinated and versatile whole-body motions.
The RL policy overcomes the widely adopted virtual-hitting-plane constraints in robotic TT, and simultaneously generates joint reference motions for arm swing and footwork.
To enable effective training and deployment, a learning-based ball-hitting predictor was designed to augment the actor policy, and dense rewards were constructed using physics-based prediction.
Simulation and hardware experiments confirmed that the unified RL policy enables the emergence of two-dimensional footwork, a key physical skill in human athletic TT. Ablation studies validated that both the learned predictor and the predictive reward design are critical for effective end-to-end learning. The learned RL policy transferred zero-shot to the 23-DoF Booster T1 humanoid robot, and achieved versatile motions and competitive returns for diverse incoming ball motions.

 Our future work will explore dexterous arm designs beyond the four-joint arm used in this study and leverage human demonstration data \cite{AMASS}, aiming to expand the repertoire of hand skills such as backhand motions. Human TT experts rely heavily on arm dexterity, with a total of 7 DoFs each arm including 3 DoFs from the wrist. In contrast, the Booster T1 arm used in this study only has 4 DoFs, restricting behavior emergence during RL training. We also plan to employ curriculum learning \cite{10.1145/1553374.1553380} to progressively acquire core TT skills, building upon recent studies on human TT stroke training \cite{ma_avattar_2024}.

\vspace{5mm}
\section*{Acknowledgment}
We would like to thank Peng Xu (Google DeepMind), Fan Shi (National University of Singapore), Zhi Su, and Zhigang Sun for their insightful discussions, and Jiaofeng Liao and Yen-Jen Wang for their valuable technical input. We also appreciate Zenan Zhu’s assistance in preparing the figures and videos, and Archer Lin’s contributions to the experimental work. Hardware support from Booster Robotics is gratefully acknowledged. Wenjing Li is supported by the Lilian Gilbreth Postdoctoral Fellowship. 
This work used Jetstream2 at Indiana University through allocation MCH250061 from the Advanced Cyberinfrastructure Coordination Ecosystem: Services \& Support (ACCESS) program, which is supported by U.S. National Science Foundation grants \#2138259, \#2138286, \#2138307, \#2137603, and \#2138296.
This material is based upon work partly supported by the National Science Foundation under
Grants CMMI-1934280 and CMMI-2046562. Any opinions, findings and
conclusions or recommendations expressed in this material are those of the
author(s) and do not necessarily reflect the views of the National Science
Foundation.
\vspace{5mm}

\bibliographystyle{IEEEtran}
\bibliography{root_ICRA26_Final}

\end{document}